# Reinforcement Learning with Frontier-Based Exploration via Autonomous Environment

Kenji Leong
Singapore Institute of Technology (SIT)
Singapore
2000553@sit.singaporetech.edu.sg

*Abstract* - Active Simultaneous Localisation and Mapping (SLAM) is a critical problem in autonomous robotics, enabling robots to navigate to new regions while building an accurate model of their surroundings. Visual SLAM is a popular technique that uses virtual elements to enhance the experience. However, existing frontier-based exploration strategies can lead to a non-optimal path in scenarios where there are multiple frontiers with similar distance. This issue can impact the efficiency and accuracy of Visual SLAM, which is crucial for a wide range of robotic applications, such as search and rescue, exploration, and mapping.

To address this issue, this research combines both an existing Visual-Graph SLAM known as ExploreORB with reinforcement learning. The proposed algorithm allows the robot to learn and optimize exploration routes through a reward-based system to create an accurate map of the environment with proper frontier selection. Frontier-based exploration is used to detect unexplored areas, while reinforcement learning optimizes the robot's movement by assigning rewards for optimal frontier points. Graph SLAM is then used to integrate the robot's sensory data and build an accurate map of the environment. The proposed algorithm aims to improve the efficiency and accuracy of ExploreORB by optimizing the exploration process of frontiers to build a more accurate map. To evaluate the effectiveness of the proposed approach, experiments will be conducted in various virtual environments using Gazebo, a robot simulation software. Results of these experiments will be compared with existing methods to demonstrate the potential of the proposed approach as an optimal solution for SLAM in autonomous robotics. In summary, this research proposes a novel algorithm that combines frontier-based exploration, reinforcement learning, and ExploreORB, making it a valuable solution for a wide range of robotic applications during autonomous navigation.

# 1. INTRODUCTION

A variety of applications, including transportation and logistics, rescue and disaster situations, planetary exploration, and other mobile robotic operations are examples of how SLAM could be deployed. Map learning is therefore a crucial challenge for mobile robots with multiple different approaches [1] that have been extensively researched in recent years.

Map learning mainly consists of both passive and active [2] approaches. Passive approach utilises the currently perceived sensory perception of the robot using its sensors to construct a map of the environment. On the other hand, an active approach plans the trajectory of the surroundings to guide the robot through a dynamic environment where the robot can simultaneously explore the environment on its own while learning about the environment. Thus, active mapping is also called autonomous exploration because it is an essential key requirement that provides the mobile robot with the ability to build a precise and accurate model of its surroundings without human intervention [3] when learning new settings.

Furthermore, to ensure proper autonomous navigation of mobile robots to function correctly, usually a map or other visual representation will be required for the robot to locate itself and perform the task assigned by the user. In the best-case scenario, the robots will be provided a priori map including all the necessary data of the environments such as obstacle's location before autonomous navigation. However, a priori map is usually not provided for active SLAM, which means that the robot will lack knowledge of the environment and it will need to create the representation on its own via exploratory techniques such as frontier-based exploration. Hence, to navigate effectively and efficiently in an unknown environment, the user is required to provide a priori map of the area beforehand for efficient autonomous navigation.

To facilitate safe planning and decision making in robotics applications, robots must have access to a consistent representation of the surrounding environment. A robot needs to be able to construct a map of its surroundings, locate itself on it, and control its own movements. Active Simultaneous Localization and Mapping (SLAM) refers to the collaborative resolution of these three key factors in mobile robots, with the ultimate objective of producing the most accurate and complete model of an unknown environment.

Traditional approaches to solve active SLAM vary in shapes and forms and can be broken down into three fundamental stages:
1) Goal Identification: limiting the number of potential destinations is often necessary to establish a finite and usually low-dimensional collection of borders between areas that are known and unknown, which are called frontiers.
2) Utility Computation: Identify a set of optimal actions, the expected uncertainty between two variables such as the robot's location and the environment map are analysed, and this can be achieved by using theory of optimal experimental design (TOED) to estimate a candidate set of actions for frontier selection.
3) Selection and execution of optimal action: determine the most valuable destination and navigate towards it based on the estimated utility.

Gazebo is a 3D robotics simulator that allows simulation of real-world physics in a high-fidelity environment [4]. For example, different types of obstacles such as glass, chairs or rooms can be simulated to determine if the robot is able to avoid these obstacles and explore these locations. Besides, by assessing factors such as the size of the rooms or types of obstacles present during frontier exploration, further optimization of the algorithms can be improved. Hence, Gazebo is an efficient and effective approach to save time and cost because it provides features that allow replication of real-world environments to determine the types of obstacles that will contribute to poor mapping results.

## 1.1 Contributions
All contribution is in the code release of ExplORB-SLAM-RL repository for ROS Noetic, representing a complete framework to showcase the improvement of reinforcement learning with ExploreORB with the aim for reproducible research in this field.

# 2. BACKGROUND
Background section will provide a breakdown of the fundamental knowledge of the key concepts required for this research.

## 2.1 ORB_SLAM2
ORB_SLAM2 (Oriented FAST and Rotated BRIEF-Simultaneous Localization and Mapping 2) [5] is a real-time visual SLAM (Simultaneous Localization and Mapping) system that enables mobile robots and other devices to navigate and create maps of their environment using devices such as camera to detect feature points. ORB_SLAM2 is built upon the original ORB_SLAM1 system [6] and includes several improvements such as the ability to handle larger maps, support for monocular, stereo, RGB-D cameras, and improved tracking and mapping accuracy. ORB_SLAM2 is a popular approach in robotics and computer vision research and has been found in widespread of diverse real-life situations, such as indoor and outdoor environments, drones, and autonomous vehicles, yielding successful outcomes.

## 2.2 Visual SLAM
Visual SLAM [7] is a technique used in robotics that enables a robot to create a map of an unknown environment while simultaneously estimating its own position within that environment through using visual data from a camera attached to the robot and algorithms such as ORB_SLAM2 to interpret the images captured to create a map of the environment. Visual SLAM is a complex process that involves several steps, including feature detection and matching, camera calibration, 3D points cloud generation, and pose estimation. Feature detection and matching are used to identify and track distinct features in the environment, such as corners, edges, or landmarks. These features will then be used to create a map of the environment, and the robot's position is estimated by comparing the current image to previous images and their corresponding maps. Process will be repeated until the end of exploration.

**2.3 Graph SLAM**
Graph SLAM [8] is a popular approach for solving the SLAM problem, where the environment is represented as a graph consisting of nodes representing robot poses and landmarks, and edges representing measurements of distance between nodes. The goal is to estimate the best configuration of the graph, which corresponds to the most likely values of robot poses and landmark positions given the measurements taken. Cost function is often represented as a nonlinear function of the robot poses and landmark positions a various optimization can be done using techniques such as least-squares, gradient descent, or nonlinear least-squares to reduce errors encountered.

**2.4 Theory of Optimal Design (TOED)**
TOED also known as optimal experimental design is a branch of statistics that deals with the design of experiments or observations to improve the efficiency and precision of statistical inference [9]. In robotics, optimal design is particularly relevant in the context of sensors placement. The goal is to determine the best locations to place sensors on a robot to maximize the amount of information obtained about the environment. Optimal design can also be used in the context of active learning and exploration, where the robot selects the most informative actions to perform and learn about the environment in an efficient manner. Hence, TOED plays a crucial role in robotics for designing efficient and effective SLAM approaches through conducting various experiments.

**2.5 Reinforcement Learning**
Reinforcement learning is a machine learning technique that involves an agent learning to make decisions in an environment through interaction and receiving feedback in the form of rewards or punishments. Agent takes actions based on the current state of the environment, and the environment responds by transitioning to a new state by providing feedback in the form of a reward signal. The goal of the agent is to learn a policy that maximizes the expected cumulative reward over time. Reinforcement learning has been successfully applied in various domains, including robotics, gaming, and recommendation systems. In robotics, reinforcement learning can be used to train agents to perform tasks such as grasping objects, navigation, and manipulation. By providing a reward signal that corresponds to the desired behaviour, agent will improve its decision-making process leading to successful task completion.

**3. PROBLEM FORMULATION**
A major issue with autonomous exploration is to determine navigation of the mobile robots during the map learning process when provided with the current knowledge of the environment to retrieve the most amount of information. As a result, it is crucial to have an efficient and effective exploration strategy that provides an optimal path that takes relatively less time when exploring the simulated Gazebo environment. Furthermore, frontier exploration is a technique that seeks to expand the known space over time by continuously visiting these boundaries known as frontier when performing autonomous navigation. Below will describe the key issues faced during frontier exploration of ExploreORB.

Firstly, frontier points are randomly generated, and selection of frontier is based on the highest information gain. Two main launch files will be required to start the autonomous exploration, which contains the use of RVIZ (ROS Visualisation) that generates a simulated Gazebo environment and initialise the decision maker process to starts the autonomous agent for exploration. For example, at the start of exploration, frontier points are randomly generated with each centroid represented as green tiles created from each frontier points cluster because centroids help in simplifying the representation of the frontiers and provides a clear target for the robot to navigate toward the goal for comprehensive coverage of the environment.

However, since frontier points are not being saved or stored either locally or on the cloud. The robots will have no knowledge which centroids it has visited previously for every iteration, resulting in additional computation and exploration time to generate a path with the highest information gain. For example, once the robot has reached a centroid, when calculating the hallucination graph to determine which centroid goal to navigate to, it will re-calculate the information gain of the previous centroids constantly. Besides, depending on the size of the environment, since centroids are constantly generated, there is a need to store and filter these large amounts of centroids containing their x and y coordinate and information gain respectively. For example, within a large environment such as a bookstore, the number of centroids generated will be higher as compared to a smaller environment such as a house.

Another challenge of ExploreORB will be the tweaking of the various parameters used for robot movement such as speed, turning velocity and acceleration limit. For example, if the robot were to move at a high speed, detection of features points may be lost due to hardware capabilities, which will result in re-localisation of the robot. Furthermore, loop closure threshold includes another parameter known as "frustum culling", where a higher frustum distance will require additional computation power to process each map point and path planning from ORB_SLAM2 and hallucinated path. Hence, multiple tests are required to select the best configuration for different environment.

Therefore, key issues of ExploreORB highlighted will be, firstly, will be frontier points not being saved or stored, resulting in every iteration of the exploration the robot will not have a dedicated or fixed path when performing autonomous navigation because the robot does not possess the prior knowledge of the frontiers it has visited before. Besides, selecting the best configuration possible through adjusting the various parameters is important because for reinforcement learning to perform optimally, high map coverage is required through visiting centroids that contains most of the landmarks in the environment.

## 4. MOTIVATION

The motivation of this project is to improve the frontier autonomous exploration process by selecting the proper frontiers when performing mapping because in a real-world environment the size of the venues such as stadium may be very large, and it is inefficient to perform mapping of every location manually. For example, in a manual process, the user will have to be present at the location and a remote control will be used to navigate the robot to each mapping location with the use of the LIDAR sensors to scan the surrounding environment.

However, if the obstacles do not contain many unique features points such as corners or edges, it will pose a difficulty for the LIDAR sensors to identify these obstacles because features points are required for obstacle detection. Hence, resulting in a rigorous and time-consuming process for the user when performing mapping manually because the robot the robot will have to move at a slow speed to every corner of the map to ensure enough time is given for the features points to be captured and updated by the LIDAR sensor.

Therefore, reinforcement learning is implemented at the end of each frontier exploration to store all the frontiers that provides the highest number of information gain for every iteration of exploration. This would include deploying an agent using a trial-and-error methods to achieve a goal in an uncertain or complex environment. The process is performed using a reward-based system [10] where the user will determine when to allocate rewards such as points for each correct or incorrect action performed by the robot. For example, points will be rewarded for every correct action when navigation to the correct goal, while deducting a point for any dangerous actions such as speeding or hitting an obstacle.

## 5. PROJECT OBJECTIVES

Objective of this report is to provide exploration tactics for the intricate surroundings that are frequently seen in real-world environments such as houses or bookstores. Strategy is centred on the identification of frontiers-based regions that stand between recognized open space and uncharted territory. A good exploration strategy includes generating a complete or nearly complete blueprint of the designated real-world environment within a reasonable amount of time.

In the problem statement, since the robot is unable to save the frontiers points it has visited before, additional computation processing time is required to visit these centroids which is inefficient. As a result, Q-learning is introduced as a model-free reinforcement learning method that allows the agent to train and perform optimally by learning the best policies from a fixed amount of robot states. However, due to the unknown number of states required for various environments due to the scale and size, deep reinforcement learning methods such as Deep Q networks (DQN) is proposed that uses a neural network for robot state optimization because it can store a continuous amount of robot states regardless of the size of the environment. To explore the effectiveness of deep reinforcement learning, four variations of DQN will be introduced: Deep Q Network, Double Deep Q Network, Dueling Deep Q Network, and Double Dueling Deep Q Network. Metrics such as execution time, map coverage, and completion rate enable a comprehensive evaluation of the various DQN models to identify which model is the best for each environment.

Optimizing robot states through deep reinforcement learning offers several advantages over storing them on a hard disk drive. One key benefit is efficient storage utilization, as saving the complete robot states, including position, orientation, and sensor measurements, would require substantial storage space. With the addition of extra sensors, this data volume could become even larger. Alternatively, by storing only the relevant observations, such as position and orientation, and employing a neural network to forecast the optimal frontier to explore, storage space can be conserved. Moreover, employing reinforcement learning models for optimizing robot states allows for portability. Once the training process is complete, these models can be saved on the cloud, making them accessible from anywhere. This enhances flexibility and ease of deployment, enabling the robots to leverage the trained models efficiently.

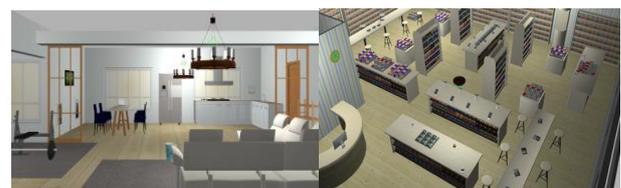

*Fig 1. AWS House (Left) and Bookstore (Right)*

Two Gazebo environments such as aws_house and aws_bookstore generated from Amazon Web Service (AWS) will be used for comparing between the four variations of DQN models mentioned above to identify the optimal algorithm for each environment. Robot states will contain the observation recorded from the sensor's measurements such as IMU (Inertial Measurement Unit), odometry and LIDAR to determine the current state of the robot, while the reward will be the information derived from the weighted hallucinated graph for each frontier for ExploreORB.

Key contributions can be summed up as follows:
- Reinforcement Learning:
    - Development of the algorithms of the

- four various of the proposed Deep Q networks.
  - Comparison of the four DQN models with metric evaluation of execution time, coverage and completion rate.
  - Experiments of algorithms in two Gazebo environments such as aws_house and aws_warehouse
- Optimal Configurations:
  - Identify the optimal parameters to be used for efficient and effective autonomous path planning.

**5.1 Breakdown of Report Format**

Breakdown of the report format is described as follows. In Section 6, a literature review of relevant research of past and current exploratory tactics active SLAM and frontier exploration techniques. In Section 7 and 8, explaining and describing the various stages of ExploreORB and deep reinforcement learning methods. In Section 9, the proposed ideas of the implementation of each DQN models. In Section 10, results and findings will be conducted to compare the results before and after implementation of deep reinforcement learning. Lastly, Section 11, will consist of a conclusion to identify optimal DQN model for each environment and limitations of the current implementations.

**6. LITERATURE REVIEW**

Below will provide a literature review to summarize and evaluate the history of active, graph and visual SLAM to help understand previous work and the need for this study.

**6.1 Active SLAM**

For a robot to operate autonomously, primarily it needs to form a model of the surrounding environment which includes localization and mapping [11] to estimate the position of the robot on the map. SLAM (Simultaneous Localization and Mapping) refers to the problem of incrementally building the map of the environment while locating the robot within it [12] and is a passive method that is not concerned with the navigation process.

**Early Development of SLAM (1980-1990)**

The foundation of SLAM emerged in the 1980s with the development of probabilistic methods for robotics and computer vision such as the Kalman filter. Kalman filter [13] is a widely used robotics algorithm to estimate the robot state such as the position orientation from various odometry. Firstly, wheel odometry [14] estimates the robot position through measurement of the rotation of the wheel and motor. Next, IMU combines both accelerometers and gyroscopes to measure linear and angular velocities to estimate the robot position and orientation. Lastly, LIDAR odometry [15] to estimate robot motion through comparison of consecutive point clouds generated from the LIDAR sensor.

**Emergence of Active SLAM (1990 - 2000)**

Active SLAM started gaining attention as researchers began exploring various exploration techniques by actively controlling the robot actions. This includes the introduction of information-theoretic approaches which involve using concepts from information theory for the robot to make and plan decisions by quantifying the information gain between various components of a robotic system such as sensors, environment, and state [16]. Information theory can be broken down into four steps as described below:

1) Sensor Fusion [17] - Optimize fusion of data from multiple sensors by weighting the contribution of each sensor through the amount of information it provides.
2) Sensor Placement [18] - Determine optimal sensor placements that maximize information gain and minimize uncertainty.
3) Task Allocation - Assign tasks to robots based on their functional capabilities, available and expected information gained from completing the tasks assigned.
4) Motion Planning [19] - Selecting optimal robot trajectories to maximise information gain to reduce uncertainty in the robot state.

**Advances in Active SLAM (2000 - present)**

Through the years, researchers continue to develop and improve various SLAM techniques and two extensive surveys have been conducted between 2016 and 2021 by Cadena et al [20] and Lluvia et al. [21] to identify the impacts of types of sensors, environments, and utility function for different SLAM approaches.

**EKF (Extended Kalman Filter)**

EKF [22] is an extension of Kalman Filter used in a non-linear system. Linear system principles include both superposition and homogeneity. Superposition refers to how the system handles summing of multiple inputs simultaneously where the output of the result can be calculated by splitting the inputs separately. Homogeneity refers to the scaling of the system based on the number of inputs which are directly proportional. However, a non-linear system does not follow the principle from a linear system of superposition and homogeneity where the relationship between both input and output are not directly proportional as they often exhibit complex behaviours which results in more calculation and computation power. Real-world systems are often non-linear relationships between variables are mostly not directly proportional and often involve complex interaction. For example, in robotics, since multiple components are interconnected, when a component such as a transistor or battery fails, there are multiple reasons that result in the failure of a component that require multiple examinations to identify the cause.

**RBPF (Rao-Blackwellized Particle Filter)**

RBPF was introduced by Murphy [23] which uses a particle filter to solve the problem of learning a grid-based map. Particle filter [24] contains the actual state of the robot or map and each of the states will be assigned a weight proportional to the probability of the state occurring. Prediction of robot motion [25] can be done by incorporating new sensor measurement retrieved from various robot odometry and observing the difference shown in the observation model as it defines the relationship between state of the system and sensor measurement. By performing resampling through updating the weights of newly generated particles, will provide a more informed and accurate information of the robot pose and with relation to the map of the environment. However, particle filters require high computational power due to the high dimensional state spaces required to store and process each particle and it is commonly used in a nonlinear system.

**Fast SLAM**
Fast SLAM is built on top of RBPF to solve the problem of active SLAM problems through efficient estimation of the robot trajectory and locations of landmarks in the environment. Two iterations of Fast SLAM have been researched which are Fast SLAM 1.0 and 2.0. Fast SLAM 1.0 [26] key features include using particle filters to estimate robot trajectory where each particle will represent a possible path that the robot could take. Under each particle, a separate filter was implemented to estimate the position of the obstacles within the environment that assumes known data association from measurement retrieved from the robot sensors such as LIDAR, camera, or sonar. However, this could lead to inaccuracies because often data association is often incorrect due to noise from the sensors, returning incorrect data measurement. Factors affecting data association includes ambiguity where sensor is unable to uniquely identify a landmark or feature, followed by cluttering where multiple objects within proximity of the sensor may cause confusion. Next, occlusion occurs when some objects may be fully obscured or blocked by another object, resulting in poor identification of obstacles. Lastly, sensor noise arises from various signal processing and transmission that results in poor transmission of data.

Fast SLAM 2.0 [27] was then developed as an improved version of Fast SLAM 1.0 where it incorporated data association uncertainty within its estimation process to ensure robustness and accuracy. Key differences include after landmark estimation using a simple filter from the sensors. Fast SLAM 2.0 will not assume perfect data association by incorporating multiple data association hypotheses for each sensor measurement. MHT (multiple hypotheses tracking) is a method used for multiple data association. Firstly, it will generate a set of possible hypotheses, followed by performing hypothesis evaluation and pruning, through assigning a score for each hypothesis using the consistency and weights of the sensor measurements. For example, when a robot captures an obstacle with two measurements, M1 and M2 but is unable to identify which measurement corresponds to which object. As a result, two hypotheses such as Hypotheses 1: "M1 associated with object A and M2 associated with object B" and Hypotheses 2: "M1 associated with object B and M2 associated with object A." Once hypotheses are created, evaluation of likelihood will be done to update the state estimate of the weighted combination to select the correct hypotheses. Hence, by maintaining and updating multiple hypotheses will help solve uncertainties and ambiguities of the sensor measurements through providing accurate object tracking.

**G-Mapping (Grid-based Fast SLAM)**
G-Mapping is implemented as an alternation variation of Fast SLAM known as Grid-based Fast SLAM [28]. Firstly, particle filtering is done to estimate robot trajectory and each particle will represent a robot path that will be mapped onto an occupancy grid. Occupancy grid is 2-dimensional representation where each cell stores the probability value to determine if an obstacle is present. For example, 0 represents free region and 1 represent occupied cell. Once particle filtering is completed, other steps such as robot motion, sensor measurements and resampling will be processed and using the Fast SLAM algorithms to select the particle that provides the highest weight that corresponds to the most accurate hypothesis. Hence, G-Mapping has similar features as compared to Fast SLAM with the addition of occupancy grid that offer several benefits such as robustness to handle noisy sensor data by measuring sensor occupancy probabilities updated in real-time from sensor measurement when mapping is performed.

**Visual SLAM**
Visual SLAM [7] is a widely proposed algorithm that uses image features captured from camera to create a map of the environment and composed of mainly three modules such as initialization, tracking and mapping. Firstly, initialization refers to calibrating the camera with a global coordinate system [29] which includes cartesian coordinates with three orthogonal axes (x, y, and z) that define a fixed set of coordinates to serve as a reference frame to specify position of the robot or object position in the environment. For example, a flat plane contains two axes (x, y) the intersection between these axes will be at origin (0, 0). Robot and goal position can be represented at coordinate (1, 2) and (3, 5) respectively. Hence, global coordinate system will provide a fixed consistent reference frame to represent position of robot and goals when surveying complex environment.

Furthermore, tracking refers to feature extraction such as corners, key point, or pixel intensity because these features allow the algorithms to perform camera and landmark estimation through comparing between consecutive frames to track obstacles or current robot position and orientation. To ensure accurate map creation, optimization techniques such as bundle adjustment from ORB_SLAM2 is used to minimize projection errors between observed image features points and estimated projection in a reconstructed 3-dimension map.

Visual SLAM has been widely used in multiple fields in computer vision, robotics, and AR (augmented reality) [30] because these technologies are suitable for camera pose estimation. For example, in AR systems, it utilizes the camera from a smartphone to identify human gestures for real-time responses with the limited computational resources of a handheld device. Furthermore, UAV (unmanned autonomous vehicles) [31] such as drone or surveillance robots also utilizes visual SLAM to be able to view streaming footage and identify the position of these vehicles on a map using various odometry that exchange data interchangeably.

**Graph SLAM**

Graph SLAM focuses on estimating the robot trajectory without reconstructing the obstacle or landmark map. Graph SLAM [32] addresses the broader SLAM graph optimization problem as it encompasses both estimation of robot poses and landmark mapping. Graph SLAM can be classified between filtering or smoothing. Filtering refers to an on-line state estimation to process measurements from sensors to update the state estimates such as robot pose and landmark map over time through various filtering techniques such as EKF and Particle Filter. As a result, filtering provides a more consistent and better representation of the robot trajectory allowing the robot to perform autonomous navigation in the environment more efficiently and can be summarized in four steps such as initialisation, motion, observation, and estimation.

For example, in the initialisation process, an empty map and an initial node at robot starting position x0 will be created. Next, in the motion steps, when the robot moves towards a new position (x1), new nodes will be created with an odometry edge linking these nodes that contains measurement of the distance travelled by the robot. Under observation, multiple landmarks may be spotted by the robot during exploration where landmarks can be classified between natural and artificial landmarks. Natural landmarks refer to features from the environment such as rocks or trees with distinct corners or key points, while artificial landmarks are human-made features such as tables or chairs. Once landmarks are identified, nodes will be created and measurement from the odometry will be taken when connecting these nodes with the edges. Lastly, estimation of the actual robot poses, and location of the landmark will be updated using filtering techniques such as EKF and the process of motion, observation and estimation will be repeated until the robot has fully explored the environment.

On the contrary, smoothing refers to estimating the full trajectory of the map and robot through processing all the available data from the sensor measurement in batches and minimizing the errors between landmark position and pose estimates through processing the entire history of all the estimates. For example, like filtering, once nodes and edges have been created after exploration of the environment, smoothing will use optimization techniques such as Gauss-Newton to solve nonlinear least square problems by adjusting the nodes position through finding the optimal configuration for the optimized node positions.

## 6.2 Frontier Exploration

Frontier exploration for autonomous robots has been an attention point that researchers have been researching for many years. Multiple studies have been done to acquire geometric precise information from Configuration space, Grid and Voronoi models. Configuration space [33] also known as C-space represents the possible configuration of all possible states such as the position and orientation of the robot or landmarks. For example, to ensure that the robot does not collide with the obstacles from the environment, obstacles are mapped into the robot C-space also known as C-obstacles. Various techniques for planning of configuration space includes potential field methods where robot and obstacles are labelled as charged particles that will repel from each other and cell decomposition method which works similarly to occupancy grid divides the C-space into multiple cells that contains a probability factor to label obstacles within the configuration map.

Grid Map [34] uses occupancy grid to represent an environment divided into a two-dimensional grid of equally sized cells that includes a probability factor which determines where each cell is free, occupied, or unknown. Furthermore, the Voronoi model [35] refers to representing obstacles as seed and generating a Voronoi diagram that consists of edges to minimize the risk of collisions. For example, in a 2-dimenstional plane that contains two seeds A and B contains with coordinates (2, 2) and (2, 5) respectively. A series of boundary lines known as Voronoi edges will be drawn in between these two seeds. As a result, the robot can follow these boundary lines when performing path planning and Voronoi diagram can be scaled up to n-dimension for complex obstacles containing many edges seen in the environment.

**Advancement in Frontier Exploration**
*Yamauchi*
Yamauchi [36] built on the idea of frontier-based exploration through the concept of gaining the most information of the environment by moving through between open space and uncharted or unexplored territory. Open space refers to environments that are known as "free" that contain few obstacles that the robot can move without complex path planning as obstacles are easily avoidable. Meanwhile, unexplored space refers to areas that have not been studied or map by the robot where it does not contain any knowledge of the environment and would require the use of sensors with the combinations of various active SLAM methods depending on the types of environments to perform mapping. Hence, both types of spaces will pose unique challenges for the robotics system during exploration.

Besides, when the robot reaches a frontier in the map, these points are known as "accessible" because these are contiguous spaces due to a path existing from the initial robot position. For example, every path within an unknown territory will encounter a frontier, and once the robot reaches the frontier, updating of the mapped frontier will be performed. Eventually, the robot will be able to explore all the accessible space assuming that in a perfect scenario of all sensor readings are accurate. However, sensor measurements often contain inaccurate measures resulting in the challenge of performing frontier exploration with noisy sensor data.

*Laser-Limited Sonar*
To solve noisy sensor data, Yamauchi mentioned using evidence grid [37] for Laser-Limited Sonar. Evidence grids are cells that contain a prior probability to provide an initial estimate of the occupancy level of the grid and this occupancy level will be updated when a new sensor reading from the robot LIDAR, ultrasonic or infrared is captured. Sonar sensors uses sound waves to measure and detect obstacles through measurement of the sound pulse when it is deflected from flat surface obstacles such as doors or wall. However, if it encounters a reflective obstacle such as mirror, the sound pulse would bounce away which will return incorrect measurements. Hence, instead of waiting for the sound pulse to bounce back to return the reading to the sonar sensor, a laser range finder can be incorporated with the sonar sensor. For example, if the reading of the laser is shorter than the range of the sonar reading, it assumes that the sound pulse has returned from the sonar. Although laser-limited reading is not the ideal solution, as lasers could also bounce off any glass objects, it has proven to reduce the number of uncorrected specular reflections.

*Entropy-Based*
Entropy-based frontier exploration [38] refers to reducing uncertainty of the entropy of the robot when exploring an unknown or partially explored environment. Environment is represented as occupancy grid holding status such as free, unknown, or occupied and within each grid would contain an entropy value from high to low that represents uncertainty of a cell. High entropy are areas with high information gain because it represents high uncertainty of areas containing many obstacles such as open areas. On the other hand, low entropy are areas with low information gain such as an empty room because there are fewer obstacles detected. Hence, robots will prioritize areas with high entropy because there are areas that will provide the highest information gain that helps to improve the accuracy of the map. Two types of entropy-based methods used are Shannon and Renyi entropy [39] to measure uncertainty using information theory to reduce uncertainty.

Shannon entropy is used to quantify the uncertainty or information content in the belief state, which includes the robot's pose estimation and the map. By calculating the entropy of the belief state, SLAM algorithms can assess the level of uncertainty of the current estimated robot pose. This information is crucial for decision-making processes in SLAM, such as selecting the most informative actions or measurements to reduce uncertainty. Shannon entropy provides a measure of the information content in the SLAM system, enabling efficient exploration and data acquisition to improve the accuracy and robustness of the estimation process by selecting actions that minimize the expected entropy to maximize the information gain, the SLAM system can efficiently explore the environment and gather data that reduces uncertainty.

Renyi entropy is a generalized measure of uncertainty or information content, extending Shannon entropy. It introduces an order parameter, α, allowing for different emphasis on rare or common events. Renyi entropy quantifies the information content in a dataset based on the probabilities of outcomes, with α controlling the weight. The order parameter α controls the emphasis on different probabilities in the calculation. For example, as α approaches 1, Renyi entropy converges to Shannon entropy and when α deviates from 1, the Renyi entropy places more emphasis on either rare events (α > 1) or common events (α < 1). Renyi entropy enables tailored sensitivity to different probabilities, providing a versatile framework for measuring information content in the dataset.

*Gaussian Processes Occupancy Maps (GPOM)*
In the problems of frontier exploration, Yamauchi proposed frontiers extracted from an occupancy or evidence grid map. GPOM [28] is an algorithm that improved Yamauchi frontier exploration by representing each grid as a continuous function that allows the robots to estimate not just occupancy probabilities but also the uncertainties that correlates to the confidence level during exploration. Besides, GPOM also handles sparse data handling when handling incomplete or limited sensor data which often happens in real-world cases. For example, LIDAR sensor may only be able to detect only a few sets of nearby obstacles due to its limited range. As a result, the occupancy map is unable to fully determine if the object is an obstacle. Hence, when there is no data available, GPOM will provide an estimate of the uncertainties for regions that have low amounts of data. However, GPOM is very computationally expensive due to multiple algorithms used to handle each grid data, which may not be suitable for low-powered devices due to the computation power required.

## 6. METHODOLGY
In this section, it will describe the implementation of the training and testing scripts of four variations of deep reinforcement methods such as DQN (Deep Q network), DDQN (Double Deep Q network), Dueling DQN (Dueling Deep Q network) and Dueling DDQN (Dueling Double Dueling Deep Q network). Flowcharts will be provided to illustrate the flow of each process.

### 6.1 Deep Reinforcement Learning
Below will describe the software architecture, flowchart and implementation of each deep reinforcement learning method used.

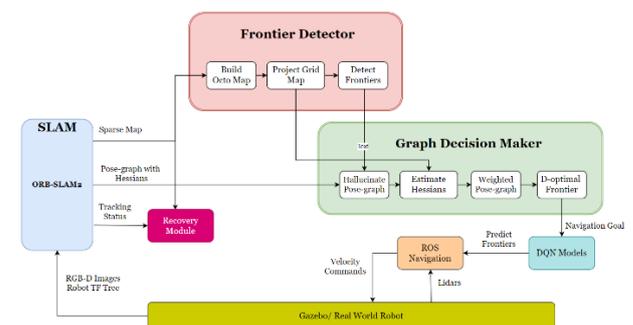

*Flowchart – RL Model with ExploreORB*

The flowchart illustrates the software architecture for integrating reinforcement learning into ExploreORB. This integration involves several steps. First, the D-optimal frontiers are generated for each centroid and information gain record. These generated inputs, along with the robot's current orientation and position, are then utilized as input for the DQN models to predict the centroids that the robot has previously visited based on previous training dataset. Lastly, ROS Navigation is employed to transmit navigation commands to the robot, guiding it towards the predicted centroids. Hence, the process will repeat itself until the robot has completed the exploration of all centroids in the Gazebo environment.

### 6.1.1 Flowchart

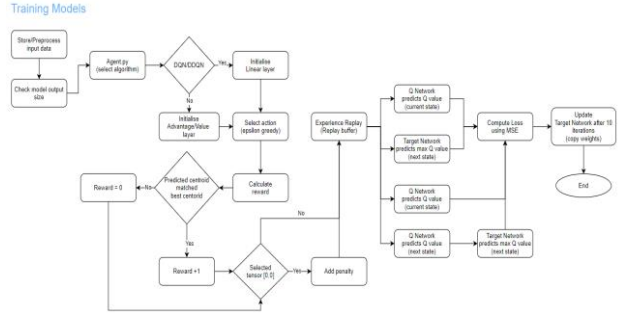

*Flowchart – Training Model*

Before training the models, preprocessing of data, and identifying the maximum number of centroids is required to fit various environments. "Agent.py" will be a handler that contains initialisation of the four variations of DQN models and each action will be selected based on epsilon greedy followed by calculating the rewards if the predicted centroid matches the best centroid with penalty being added if DQN model selected centroid with tensor [0, 0]. During the training process, a total of 100 epochs or episode will be used, where epochs refer to the numbers of time the entire training dataset is being trained. Weights of the Q network will be copied to the target network with an interval of every 10 epochs, to provide stability for the target network. Lastly, Q-values of each robot state will be stored in the experience replay or replay buffer to learn various patterns and predict predicted centroid.

### 6.1.2 Prepare Inputs

To prepare inputs for the model, conversion of NumPy array to PyTorch tensor of the robot position, orientation, centroid and information gain record are required because it has built-in features such as automatic differentiation for matrix multiplication in a neural network.

The robot state of will utilise "torch.cat" that will concatenate and flatten into one tensor that contains this robot observation, with an input size of length 36 with robot position containing 2 values (x, y), robot orientation containing 4 values (x, y, z and w), centroid record with 20 values (list of 10 pairs of (x, y)) and information gain with (10) values where the index of each pair of centroid record correlate the index of the information gain records. .

$$network_{output} = centroid\_record.shape[0]$$
(1)

Besides, the output size will a length of 10 because there will be 10 actions or centroids that the model can choose from. In the equation, the "shape" function is used to return the number of columns in the tensor.

### 6.1.3 Initialise Model

A handler will initialise the models, with the necessary parameters based on the selected algorithm selected by the user. Below will describe the parameters.

| Parameters | Description |
|---|---|
| gazebo env | Select environments "aws_house" or "aws_bookstore". |
| algo | Select algorithm "dqn", "ddqn", "dueling_dqn", "dueling_ddqn". |
| repeat count | Number of times the agent explores the environment. |
| gamma | Determines the importance given to future rewards in the decision-making process. A higher gamma value places more emphasis on long-term rewards, while a lower value prioritises immediate rewards. For example, 0.9 suggests agents prioritise future rewards. |
| learning rate | Control the rate model learns from the training data. Higher value will lead to higher convergence. Range is between 0.1 to 0.0001. |
| epsilon | Start epsilon at initial value of 1. |
| epsilon min | Minimum epsilon value will be used during exploration. For example, 0.1 represents 10% exploration over exploitation. |
| epsilon decay | Rate of epsilon decreases where higher value means the agent relies more on learned policy because it will explore less. |
| epochs | Number of times the entire dataset is passed through. |
| save interval | Number of epochs before updating the weights in the target network. |
| batch size | Number of rows processed at single time. |
| penalty | Penalty when model selects a centroid with tensor [0, 0]. |

*Table 1: Parameters*

### 6.1.4 Replay Buffer

Replay buffer is a data structure used in reinforcement learning to store and sample experiences. It acts as a memory for an agent by storing transitions of state, action, reward, next state, and episode termination flag.

$$self.replay\_buffer = ReplayBuffer\ (10000) \quad (2)$$

A class of "Replay Buffer" will be created and contains methods such as "push" and "sample" that will be utilise by all the DQN models. The replay buffer maintains a fixed-size buffer, allowing new experiences to replace old ones when the buffer is full. During training, experiences are randomly sampled from the replay buffer to create training batches enabling an agent to learn from past experiences and improve its policy over time.

**9.1.5 Action Selection**

Action selection with epsilon-greedy is a common strategy in reinforcement learning. It involves selecting an action based on a trade-off between exploration and exploitation. The agent selects a random action using an epsilon greedy policy, allowing for exploration and exploitation.

$$Epsilon = if\ self.epsilon > self.epsilon\_min$$
$$self.epsilon\ *= self.epsilon\_decay \quad (3)$$

However, when the epsilon is greater than the minimum epsilon value, the exploration rate is decayed using the "epsilon decay" factor. With a probability of (1 - epsilon), the agent selects the action with the highest estimated value, exploiting the knowledge it has gained so far. By gradually decreasing epsilon over time, the agent transitions from exploration to exploitation, focusing more on the learned optimal actions. Epsilon-greedy balances the exploration of new actions and the exploitation of known good actions to improve the agent's performance.

$$Action\ = random.randint\ (0, output\_size - 1) \quad (4)$$

Firstly, if a random number generated is less than the exploration rate of the "epsilon", a random action will be chosen from the available actions which will contain an integer value range between index 0 to 9.

$$Q_{values}[0, indices] = -self.penalty \quad (5)$$

However, if the random number is greater than the epsilon value, Q-values are obtained from the Q-network "self.DQN" and any centroids containing coordinates [0, 0] will be penalised to prevent the action from selecting centroid with coordinate [0, 0].

$$Action = Q_{values}.argmax(dim = 1).item \quad (6)$$

Hence, the action with the highest $Q_{values}$ will return as the selected action because it contains the highest cumulative rewards when compared with the other actions.

**6.1.6 Rewards**

Rewards play a crucial role in reinforcement learning as they provide feedback to the agent about the desirability of its actions. In reinforcement learning, rewards are used to guide the learning process by reinforcing good actions and discouraging bad ones. Positive rewards are given for desirable outcomes, incentivizing the agent to repeat similar actions in the future. Negative rewards or penalties are assigned for undesirable outcomes, encouraging the agent to avoid those actions. Below will explain each step to compute rewards.

*Predicted Centroid*

To get the predicted centroid with the highest information gain. the function evaluates the target network on the prepared input to obtain the output.

$$Predicted\ Centroid_{index} = self.target\ network(Q_{values}.argmax(dim = 1).item) \quad (7)$$

Centroids with coordinates [0.0, 0.0] are identified and their corresponding output values are replaced with a penalty value. The function then determines the index of the centroid with the maximum output and retrieves the actual centroid coordinates from the centroid record that contains the highest information gain.

*Target Centroid*

Target centroid refers to the best centroid records from previously visited centroids. This initial target will be used as one of the criteria to penalise the agent if chosen predicted centroid that does not match with the target centroid.

$$match = torch.all(torch.eq(predicted\ centorid, target\ centroid) \quad (8)$$

In the equation above, it uses the "torch.eq" function to perform element-wise equality comparison between the `predicted_centroid` and `target_centroid` tensors. The "torch.all" function is then used to check if all the elements in the resulting tensor are `True`, indicating a match between the predicted and target centroids.

$$Reward = if\ match:$$
$$reward = 1$$
$$reward = 0 \quad (9)$$

In the equation, if a match were to return "True", reward will increase by 1 else if return "False" reward will be 0 to denote wrong selection of centroid.

*Penalty*

$$torch.all(torch.eq(predicted\ centorid, zero\ centroid) \quad (10)$$

In the equation above, "zero centroid" tensor is created that contains coordinates [0.0, 0.0], similarly if the predicted centroid were to match zero centroid, additional penalty is added to reward.

$$Reward = -self.penalty \quad (11)$$

The purpose of the penalty is to discourage the agent from selecting zero centroid because as the exploration reaches to the end, the number of centroids available will inevitably start to decrease. For example, as robot explore more centroids in the environment, the coverage will increase and after a certain period of exploration, the number of centroids available will decrease to 0 eventually. Hence, adding penalties in various sections of the codes such as calculating rewards, selection actions or predicting centroids are performed to prevent the model from selecting zero centroid because this centroid value does not provide any information gain.

### 6.1.4 DQN (Deep Q-Network)

DQN [39] is a reinforcement learning algorithm that combines Q-learning with deep neural networks. It uses a neural network called the Q network to estimate the Q-values for each action in each state. During training, experiences are stored in a replay buffer and sampled randomly to update the Q network. The target Q-values are calculated using a separate target network with delayed updates and it employs an epsilon-greedy policy to balance exploration and exploitation.

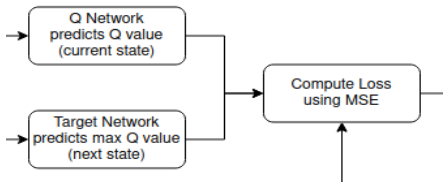

The Q-values will be calculated using the current state on the Q network "self.DQN", while the Q-values of the next states will be computed using the target network "self.target_DQN".

$$Target\ Q\ value, Q(S,A) \\ = Q(S,A) + \alpha(R + \gamma maxQ(S',a) \\ - Q(S,A)) \quad (12)$$

In the equation, to calculate the target Q-values, addition of the current Q-value state ($Q(S,A)$) with the learning rate ($\alpha$) multipy with the rewards ($R$) plus the discount factor multiplied with the maximum Q value of the next state ($\gamma maxQ(S',a)$) subtracted with the current state. The targets are obtained by adding the calculated values to the current Q-values, and then expanding them to have the same shape as the original Q-values tensor.

*Neural network*
Below displays the neural network used for DQN.

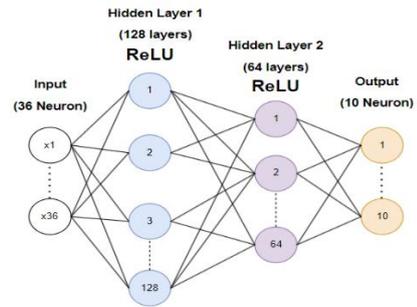

DQN neural network contains an input layer of size 36 because it contains 2 from robot position, 4 from robot orientation, 20 for each list of list inputs from centroid records and 10 from information gain records. From multiple, two hidden layers of size 128 and 64 respectively was created and in each hidden layer it uses an activation function ReLU (Rectified Linear Unit) with output layer with 10 neurons representing 10 different centroids for the agent to select from.

$$ReLU\ (x) = \max(0, x) \quad (13)$$

ReLU activation function allows mapping to model complex relationship of data by setting all negative values with minimum value of zero and $x$ will be a positive value of the Q-values because it focuses on relevant state-action pairs instead of less important Q-value such as centroid [0, 0].

$$torch.all(torch.eq(predicted\ centorid, zero\ centroid) \quad (14)$$

In the equation above, "zero centroid" tensor is created that contains coordinates [0.0, 0.0], similarly if the predicted centroid were to match zero centroid, additional penalty is added to reward.

### 6.1.5 DDQN (Double Deep Q-Network)

DDQN [40] is an extension of the DQN algorithm that addresses the overestimation bias issue present in traditional Q-learning methods that also utilises two separate neural networks such as the Q network and the target network. The Q network is used to select actions based on the current state, while the target network is used to estimate the Q-values of the next states.

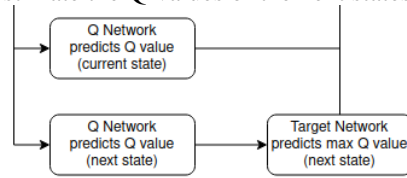

The Q-values are first computed using the online network for both the current and next states. The Q network is then utilized to select the best action in the next state based on the Q-values. The best action selected by the Q network is employed to select the corresponding Q-values from the target network. Finally, the target Q-values are computed by considering the immediate rewards, discounted maximum Q-values from the target network, and episode termination.

The purpose of using the Q network for action selection is to ensure that the agent explores and learns based on its current policy. The selected actions are then used to estimate the Q-values for the next states, which are obtained from the target network. By decoupling the action selection from the Q-value estimation, DDQN helps mitigate the overestimation bias issue that can arise in traditional Q-learning methods.

$$Target\ Q\ value, (S, A) = R + \gamma(maxQ(S', a) * (1 - dones))$$
(15)

In the equation above, the targets are calculated as the sum of the immediate rewards and the discounted gamma multiplied with the maximum Q-values for the next states by considering the episode termination. The discounted maximum Q-values are multiplied by "(1 - dones)" to apply the discount factor only when the episode is not terminated "dones = 0" ensuring that future rewards are considered when the episode continues, where "dones" is a binary value that outputs either 0 or 1.

The key difference between DDQN and DQN lies in how the maximum Q-value in the next state is calculated. In DDQN, the process involves using the Q network to predict the Q-value first, followed by passing it into the target network. Additionally, DDQN incorporates the use of the "dones" parameter, which helps identify if an episode has been terminated and this is crucial because DDQN considers scenarios where episodes may not have a clear terminal state due to complex environment where the termination of an episode cannot be easily identified such as large environment.

### 6.1.6 Dueling DQN (Dueling Deep Q Network)

Dueling DQN [41] is another extension DQN and DDQN algorithms and aims to address the challenges of learning state values and action advantages simultaneously. Both Dueling DQN and Dueling DDQN have the same network architecture, except on how the target Q-values are calculated where Dueling DQN uses a standard DQN approach while Dueling DDQN uses the standard DDQN approach as mentioned above. Like DQN and DDQN neural networks, it contains the same network architecture of 128 and 64 neuron for both hidden layers with the addition of advantage and value stream.

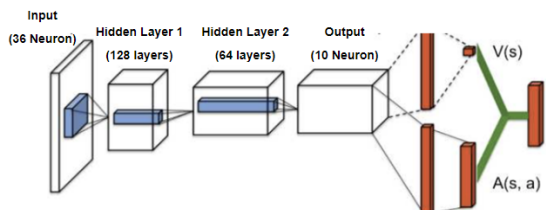

In Dueling DQN, the neural network architecture is divided into two streams such as the value stream $V(s)$ and the advantage stream $A(s, a)$. The value stream estimates the state value, which represents the expected cumulative reward starting from a given state while the advantage stream estimates the advantages of each action, which capture the additional benefit of choosing a specific action over others.

The "feature" layer is used for extracting common features from the input states consisting of a fully connected layer with 128 units with ReLU as the activation function. The "value_stream" will estimate the state value using the output from the "feature" layer through another connected layer with 64 units and the ReLU activation function will output a single value unit that represents the estimated state value, and it represents the expected return an agent can receive from a given state.

$$Target\ Q(S, A) = V(s) + \left(A(s, a) - \frac{1}{|A|} \Sigma A(s, a)\right)$$
(16)

To calculate the Q-values, start by calculating the mean advantage value across all actions for each state by subtracting the mean of the advantages along the action dimension "dim=1" from the advantages themselves. By doing so, this step helps to reduce the bias introduced by the absolute scale of the advantages. Lastly, adding the state value to the adjusted advantage allows the network to learn the relative importance of each action with respect to the state, rather than just the absolute values.

### 6.2 Decision Maker

Below will describe the steps for the training and testing process to launch the respective ROS launch files to collect sensors data to train the DQN models.

### 6.2.1 RVIZ

RVIZ is a powerful 3D visualization tool in the Robot Operating System (ROS) framework by providing a friendly interface for visualizing and analysing robot sensor data, robot models, and robot trajectories. With RVIZ, users can create interactive displays of sensor data, such as point clouds, laser scans, and camera images, to gain insights into the robot's perception of the environment. Additionally, RVIZ allows users to visualize robot models, including joints, links, and frames, enabling robot configuration, and debugging. RVIZ is highly configurable, supporting various plugins and options for customization and is widely used in robotics research, development, and simulation to visualize and understand robot behaviours and perception.

### *TEB (Time Elastic Band) and Global Planner*

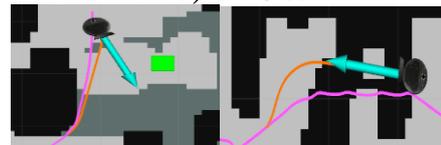

Above display both types of planners used by the robot such as TEB local planner and a Global planner displayed in colour orange and pink respectively. TEB local planner is a path planning algorithm used in ROS to generate dynamically feasible and collision-free trajectories for mobile robots by optimizing timing and curvature commonly used for smooth and efficient navigation in complex environments. Meanwhile, a global planner is a component in ROS that generates a high-level, long-term plan to guide the robot from its current location to the centroid through utilising algorithm such as Dijkstra's algorithm to find the shortest path between robot and centroid in a weighted graph. Output of a global planner is typically a sequence of waypoints or poses that the robot can follow using a local planner to navigate towards the goal.

*RGB-D Scan*
RGB-D refers to the process of capturing both colour RGB and depth information of a scene using a sensor that combines a traditional RGB camera with a depth sensor. RGB-D provides a richer representation of the environment compared to traditional RGB-only scans, as it includes information about the distances to objects in addition to their visual appearance.

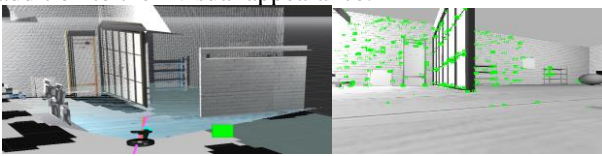

RGB-D matches the visualization based on the ORB_SLAM2 result captured from the camera in one of the areas in the aws_house environment by displaying a 3-dimensional reconstruction of the detailed point clouds to enable accurate depth measurements of the scene. By combining RGB and depth data, RGB-D scans provide a more comprehensive and detailed representation of the environment, enabling a wide range of applications that require both visual and geometric information.

*Local Cost map and LIDAR*
In robotic navigation systems, a local cost map is a representation of the immediate surroundings of a robot by providing information about obstacles, free space, and other relevant features in the local environment. The local cost map is typically generated based on sensor data, such as LIDAR sensor. LIDAR is a sensor technology that uses laser beams to measure distances to objects in the environment by providing a precise 3D point cloud data, which can be used to detect obstacles, estimate distances, to generate a representation of the local environment.

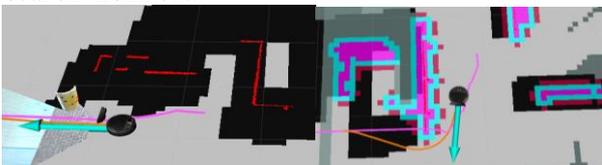

It displays the results of the LIDAR sensor when an obstacle is detected as shown on the red outer line. On the right, will be the combination of LIDAR and local costmap, colour red represents obstacles that are potentially dangerous to the robot such as a wall detected from the LIDAR sensors, blue and pink represent the outline and inner and outer layers of the obstacles. Hence, by integrating LIDAR data into the local costmap allows the robot to have an up-to-date perception of its immediate surroundings, enabling it to make informed navigation decisions and respond to dynamic changes in the environment effectively.

**6.2.1 Robot Goal**
Robot goal are centroid goals that the robot will move by providing a coordinate that contains both [x, y] location because ExploreORB uses Graph SLAM to identify various centroids and robot location.

To send the robot towards a frontier, Euler angles are used that consist of using rotational matrix to represent objects in a three-dimensional space. A 3x3 matrix contains the cartesian coordinate systems x, y, and z axis.
- x axis: horizontal direction of moving left and right from negative to a positive value.
- y axis: vertical direction, upwards and downward movement.
- z axis: depth or direction of the x-y plane.

$$Euler\ Rotation\ (R) = Rz(\emptyset) * Ry(\theta) * Rz(\varphi)$$
( 17 )

In the Euler angles equation, three rotation matrixes are computed by multiplying each matrix because each matrix represents various rotation sequence where" Roll" $Rz(\emptyset)$ represent the x-axis, "Pitch" $Ry(\theta)$ represents y-axis and "Yaw" $Rz(\varphi)$ represent z-axis. However, Euler angles has a common issue known as "Gimbal lock" when two rotational axes aligned with each other, resulting in unexpected behaviour of the robot movement.

$$Quaternion\ (Q) = (x, y, z, w)$$
( 18 )

To overcome this issue, quaternions representation is often used after the computation of the Euler angles because quaternions use a 4-dimensional space instead of three with the addition new axis "w". Calculation of quaternion requires the current and goal position as input to the Euler Rotation by setting the first two x, y axis to zero to maintain an initial orientation before updating the new quaternion value. For example, in the case of robotics quaternions can be represented:
- x (horizontal) axis: left and right direction
- y (vertical) axis: up and down direction
- z (depth) axis: forward and backward direction
- w (scaler) axis: rotation angle of the robot

Hence, by using converting the Euler angles to quaternions, it provides a more accurate representation of the robot rotation by overcoming the limitation of Euler angles that prevents a gimbal lock from occurring in a 4-dimensional space.

Besides, after prediction of the predicted centroid, to send the robot towards the predicted centroid, an input of coordinate [x, y] will be required and a Boolean value = "True" to indicate that the frontier point is valid. However, during the path planning of the hallucinated path, where if it fails to plan a path, the condition "if robot_. gestate ()! = 3" will be initialized to check if the robot's state is equal to 3. If the condition is satisfied, it means that the robot's plan has failed near its starting position.

$$Eucliden\ Distance\ d(p,q) = \sqrt{\sum_{i=1}^{n}(q_i - p_i)^2}$$

( 19 )

Euclidean distance refers to measurement of two points in map containing both coordinates (x, y). In such a case, the code calculates the Euclidean distance between the robot's current position and the initial plan position and if the distance is less than or equal to 2.0 units, it indicates that the goal has been aborted near the previous pose. To handle this situation, the code logs a warning message indicating the goal abortion and the new goal that will be sent to the robot, which is the "predicted_centroid". Finally, the "robot_. send Goal ()" function is called again to send the new goal to the robot, with the `True` argument indicating that it should clear the previous goal before sending the new one.

## 7. RESULTS AND FINDINGS
The goal of the research is to test two different Gazebo environments aws_house and aws_bookstore to compare which Deep Q network is the most optimal for environments with different sizes and obstacles.

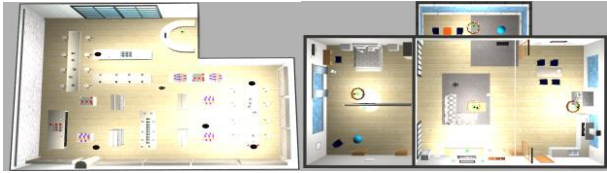

Above displays the layout of each Gazebo environment where aws_bookstore is a larger environment consisting of more obstacles and narrow areas as compared to aws_house.

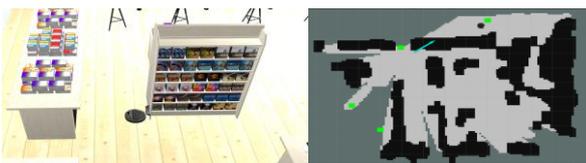

However, in the current implementation of ExploreORB, it is unable to fully map the bookstore environment as mentioned in the evaluation section of the paper [31]. As shown, if there are two bookshelves between the robot, the robot will be unable to re-localise itself back to the previous centroid and at the same time it will also lose the ORB_SLAM2 key points detection. Hence, the bookstore environment will not be used for the evaluation because for the DQN models to work, the initial training result will need to be able to map most of the environment for the model to remember the path it has visited previously for deep reinforcement learning to be optimised.

### 7.1 Evaluating DQN Models
During training for each Deep Q networks, the MSE (Mean squared error) will be calculated between the predicted and target centroid and will consist of four plots for each set of 5, 10, 15 and 20 experiments. By doing so, these plots are used to evaluate how efficient and effective each DQN model is at predicting the best centroid with the highest information gains for aws_house.

$$MSE = \frac{1}{N}\sum(Target\ Centroid - \widehat{Predicted\ Centroid})^2$$

( 20 )

MSE is a loss calculation method used to measure the discrepancy between actual target Q-values and predicted Q-values. It involves squaring the difference between these values, effectively removing any negative signs. By minimizing the MSE, the goal is to improve the decision-making process in the environment, specifically in selecting the best centroid for exploration. Hence, MSE loss minimization process aims to enhance the DQN model ability to approximate optimal Q-values, leading to more informed and effective decisions during the exploration phase.

**Deep Q Network (DQN)**

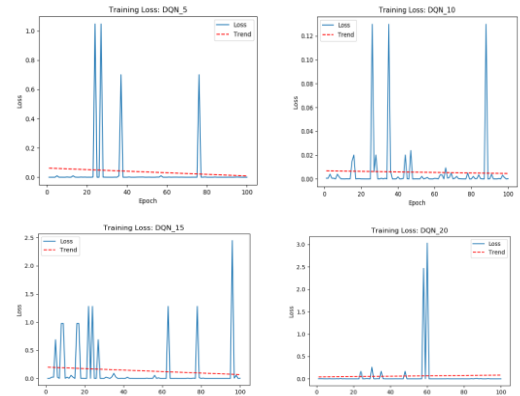

Shows the plotted result for MSE loss for DQN in all experiments for various dataset sizes, in all sets of experiment, MSE for DQN 5 ,10 and 15, the trends show that MSE has decreased as the dataset gets larger. However, in DQN 20, the MSE remains constant, and reason may include the dataset has reached a convergence point where it has successfully minimised the accuracy loss through gradient descent in MSE at the start of the few epochs. Besides, there are spikes during training of models, because for certain centroid it contains many obstacles resulting in higher information gain and if the DQN model predicts wrongly for best centroid, will result in spikes of data due to large differences between the MSE loss.

**Double Deep Q Network (DDQN)**

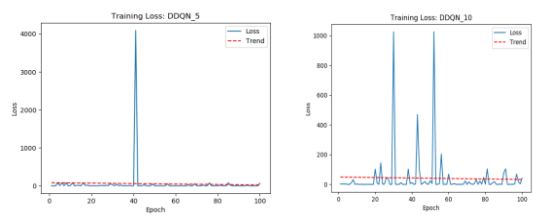

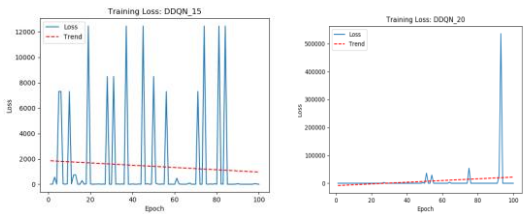

Like DQN models, once the model reaches the experiment that contains 20 rounds, the MSE loss increased and reasons may be due to overfitting where the model starts to memorise the training data too much, resulting in poor prediction for new centroid.

**Dueling Deep Q Network (Dueling DQN)**

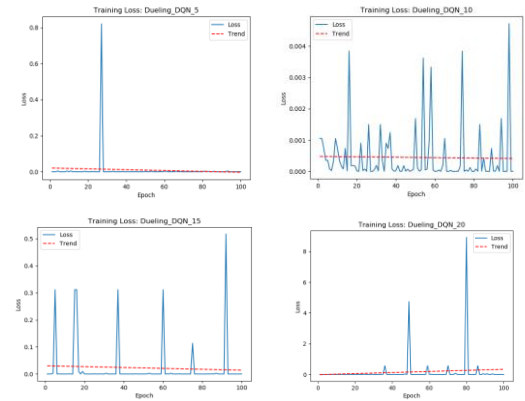

Similar for the Dueling DQN 20, there is an increase in MSE loss. Besides, in Dueling DQN 10, it has more spikes as compared to DQN and DDQN, which means the model is unable to predict the best centroid accurately as shown in the constant straight line for the average trend.

**Dueling Double Deep Q Network (Dueling DDQN)**

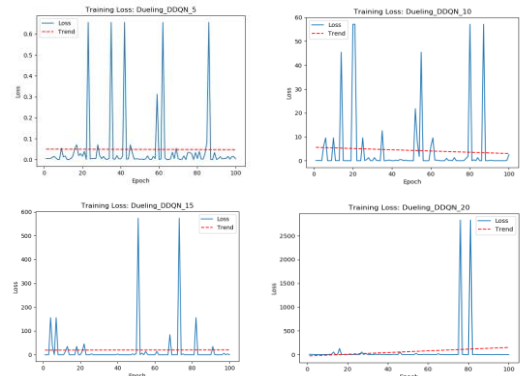

As compared to the other models, in Dueling DDQN 5, it has already reached the minimum gradient point and is reflected for Dueling DDQN 10 and 15 models with a non-noticeable downward trend. Meanwhile, Dueling DDQN 20 model, the MSE loss still increases.

**Conclusion**

In conclusion, the analysis of the plotted charts for all DQN models reveals that the average MSE loss for each model generally remains within the range of 0 to 1. This indicates that the models are improving, as evidenced by the downward trends highlighted in red. However, during the training of models on the 20th dataset for 100 epochs, there is an increase in MSE loss for most models, which may suggest overfitting of the data. Additionally, the presence of spikes in the charts demonstrates that there is no single model that consistently exhibits a downward trend for all sets of data. Therefore, it can be concluded that each DQN model has its own advantages and disadvantages, and their performance varies depending on the specific dataset and training conditions.

**7.2 Evaluating Charts**

To perform evaluation for the different Deep Q networks, metrics include comparing the average time taken, number of completed and not completed maps and the quality of the map after testing the model. Four sets of experiments were done in the aws_house Gazebo environment for 5, 10, 15 and 20 rounds to identify which model is the most optimal for this environment. Below will display the results of the DQN models where the colour blue and red represent both training and model results respectively.

**Deep Q Network (DQN)**

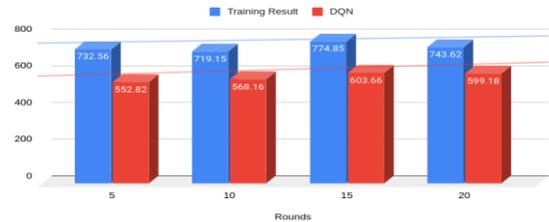

Results show that DQN performs better as it has a lower average time of around 600 seconds as compared to training results with an average time of 720 seconds.

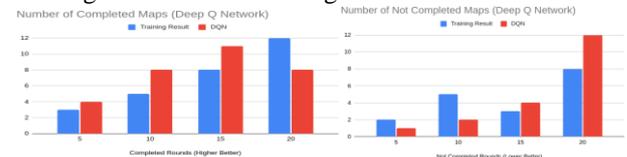

For the number of completed maps, DQN performs better than training results for 5, 10 and 15 rounds. However, during testing for a 20-round experiment, training results perform better than DQN. Similarly, the number of not completed maps for DQN is higher than the training results which means that the results of DQN models are not consistent during the experiments.

**Double Deep Q Network (DDQN)**

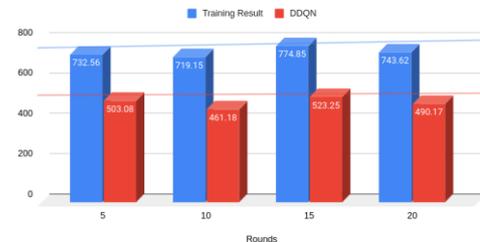

*Fig 58. Average Time Before/After DDQN*

Results show DDQN performs better as it has a lower average time of around 500 seconds as compared to training results with an average time of 720 seconds.

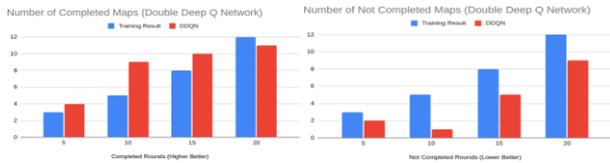

Similarly, for the number of completed maps, DDQN performs better than training results. However, during testing for the 20-round experiment, the training results still perform better than DDQN with 12 completed maps as compared to 11 but when compared to DQN, it performs slightly better than DQN as there are more completed maps.

**Dueling Deep Q Network (Dueling DQN)**

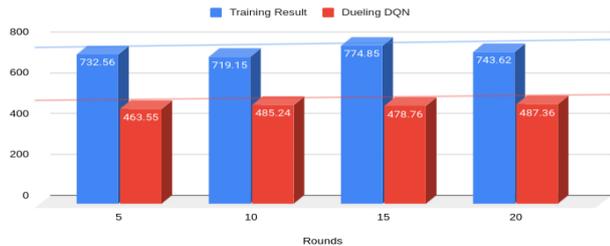

The results show that Dueling DQN performs better as it has a lower average time of around 480 seconds as compared to training results with an average time of 720 seconds.

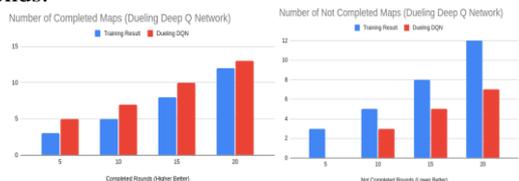

For the number of completed maps, Dueling DQN performs better than the training results for all four sets of experiments.

**Dueling Double Deep Q Network (Dueling DDQN)**

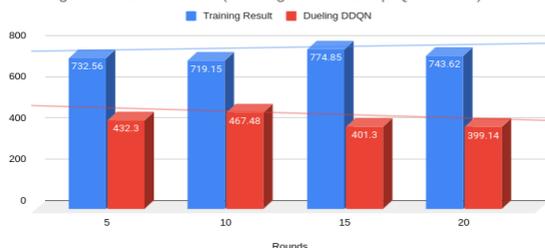

The results show that Dueling DDQN performs better as it has a lower average time of around 400 seconds as compared to training results with an average time of 720 seconds.

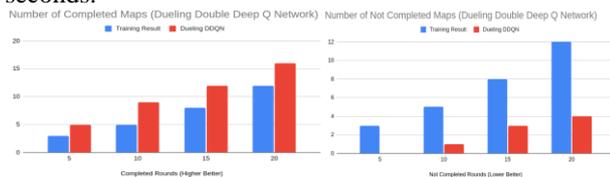

For the number of completed maps, similarly, to Dueling DQN, Dueling DDQN also performs better than the training results for all four sets of experiments.

**Comparison of all DQN Models**

Below will display the comparison between all DQN Models for the average time, number of completed and not completed maps.

**Average Time**

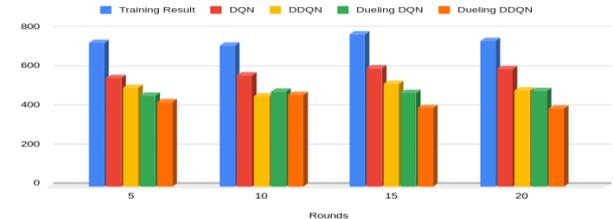

The average time for all four Deep Q networks performed better than the training results where DQN improved from 720 seconds to around 600 seconds, followed by DDQN and Dueling DQN which had another improvement to around 500 seconds. Lastly, Dueling DDQN performs the best because it has the most improvements with an average time of 400 seconds to fully complete the map of "aws_house".

**Number of Completed Maps**

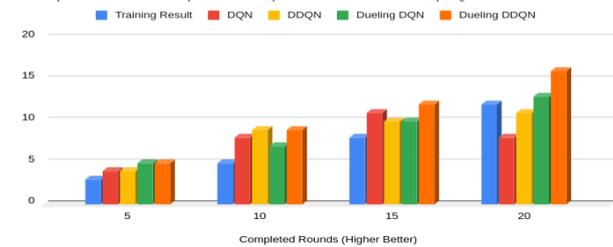

Within all four experiments, for experiments 5, 10 and 15, the number of completed maps for all the deep Q networks performed better than the training results, which is important because the higher number of completed rounds means that the robot has more records and experiences for all centroids visited. However, in the experiment for set 20, training results perform better than DQN and DDQN while Dueling DDQN performs the best out of all four experiments.

**Number of Not Completed Maps**

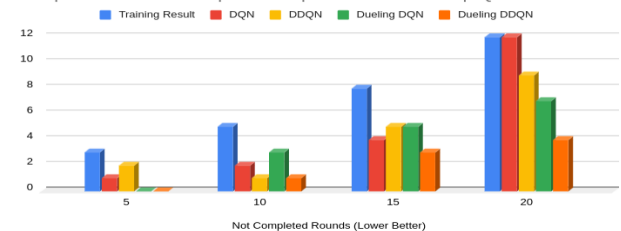

The lower the number of not completed maps is better because this allows the robot to visit more centroids so more data can be added into the model for training to ensure that the models are effective and efficient. Similarly, the results also show that Dueling DDQN performs the best for all experiments when compared with the other deep Q networks.

**Conclusion**

In conclusion, after comparing various aspects of different DQN models, it is evident that Dueling DDQN stands out as the top performer. It exhibited significant improvements in terms of average completion time, reducing exploration time by approximately 40 percent from 700 to 400 seconds. Additionally, Dueling DDQN achieved the highest number of completed maps compared to other DQN models, indicating successful and consistent mapping of the aws_house environment.

### 7.3 Evaluating Completed Maps

The evaluation of the completed maps involved the utilization of three CNN (Convolutional Neural Network) models: InceptionV3, VGG-16, and Resnet50. These models were employed to compare the best completed training image with all other completed images generated by the Deep Q networks. Through repeated testing, completed maps of each DQN models were chosen if they exhibited a cosine similarity value exceeding 75 percent because this criterion ensures that the completed DQN maps have captured most of the features present in the aws_house environment.

#### 7.3.1 InceptionV3

InceptionV3 [42] is a deep learning model used for image classification built upon the concept of "Inception modules" that employ multiple parallel convolutional layers with different filter sizes to capture features at various scales. InceptionV3 is built on a training set called ImageNet that contains 14 million annotated images of various objects and contains 48 hidden layers to classify images.

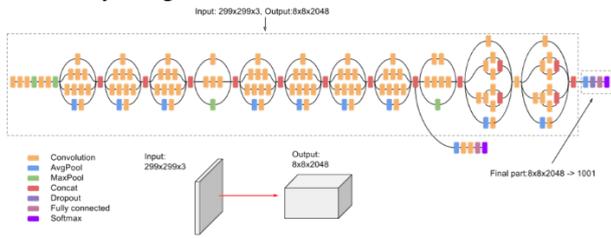

A combination of 1x1, 3x3 and 5x5 convolutional filters were used to capture various patterns within an image. Convolutional filters are matrices that extract features points within an image and depending on the quality of the images, the types of filters will affect the features detected.

*1x1 Filter*
In a 1x1 convolutional filter, with an image size of 3x3x3, it refers to 3 channels of 3x3 matrix for channels RGB (red, green, and blue). The maximum number of weights will be initialised depending on the depth size of the channels and with 3 randomise initialised.

$$Dot\ product\ (a*b) = a_x b_x + a_y b_y$$
(21)

In each channel, it will select a random weight and perform the dot product for each position starting from (0, 0) to (3, 3). In the equation above, display the formula to calculate the dot product of each channel by capturing the relationship of the input with the corresponding weights.

*3x3 Filter*
In a 3x3 filter, the convoluted image will perform by initialising a kernel with arbitrary random weights and perform element-wise multiplication. For example, if the kernel size is 3x3, depending on the stride value where stride refers the step size filter will that move the kernel horizontally from left to right. However, "Pooling" was introduced as a down sampling operation to reduce the dimension of the input features map for filters larger than 1x1. Maximum and average pooling are the two common methods to extract features from the convoluted image.

$$Output\ size = \frac{Input\ size - Pool\ size}{Stride + 1}$$
(22)

In the equation, it will determine the output size after performing the pooling operation, the dimension of the input size is 299x299x3 with pool size 2x2 and stride of 2. For example, if max pooling was used, for every 2x2 filter, it will extract the highest value in the filter. Hence, max pooling will reduce the size of the image with a final output size of 148x148 dimension.

*5x5 Filter*
In a 5x5 filter, the key difference between 3x3 filter is the increased receptive field that captures more features point in the spatial range suitable for detecting images that consist of many obstacles with many edges. For example, centroid location that contains higher information gains contains many landmarks to process. Hence, 5x5 filter captures more information with addition computation required.

*SoftMax Function*
SoftMax is an activation function used in the output layer for multi-class classification by using a probability distribution between 0 and 1.

$$Softmax\ \sigma(\vec{z}) = \frac{e^{z_i}}{\sum_{j=1}^{K} e^{z_j}}$$
(23)

In the equation, $\vec{z}$ represents the input vector, $e^{z_i}$ represents the standard exponential function that will always return a positive value above 0. For example, the initial input is positive, the exponential function will return a larger value, likewise, if input is a negative value, it will approach 0 rapidly. However, since the inputs are not in the range between 0 and 1, through normalisation $\sum_{j=1}^{K} e^{z_j}$ of number of classes $K$. As a result, the probability distribution of all the classes will add up to 1, with the highest probability as the chosen label.

*Cosine Similarity*
Cosine similarity refers to a metric that determine similarity between two vectors by calculating the cosine of the angle that will indicate the direction and proximity in a multi-dimensional space.

$$Magnitude\ ||x|| = \sqrt{a^2 + b^2}$$
(24)

In the equation, magnitude $||x||$ refers to finding the length of vector $x$ through the sum of squares of the coordinate ($a^2 + b^2$) to remove negative values and performing the square root operation $\sqrt{a^2 + b^2}$ to find the length of vector $x$.

$$Cosine\ Similarity\ (x,y) = cos(\theta) = \frac{(x.y)}{(||x|| * ||y||)} \quad (25)$$

In the equation, dot product is performed between two vectors $(x.y)$ followed by dividing the magnitude of $||x|| * ||y||$. Hence, it will return a value between -1 to 1 where -1 indicates perfect similarity of two vector pointing in the same direction, while 0 indicates no similarity with orthogonal vector and -1 indicates opposite vectors such as positive and negative values.

Both the trained and model images are resized to a dimension of 299x299 to be able feed the input as the model. The next step involves converting the image into a NumPy array, which helps in reducing memory storage space. To process one image at a time, a batch dimension is added using the "expand_dim" function. The input is pre-processed using a PyTorch class that handles additional requirements specific to InceptionV3 and extracts feature points. Finally, cosine similarity is utilized to compare two images and determine their similarity. On average, the similarity ranges from 75% to 81%.

**7.3.2 VGG-16**
VGG-16 [43] consists of 16 layers, including convolutional, pooling, and fully connected layers and it is known for its simplicity and effectiveness in image classification tasks, featuring multiple stacked 3x3 convolutional filters and max pooling layers.

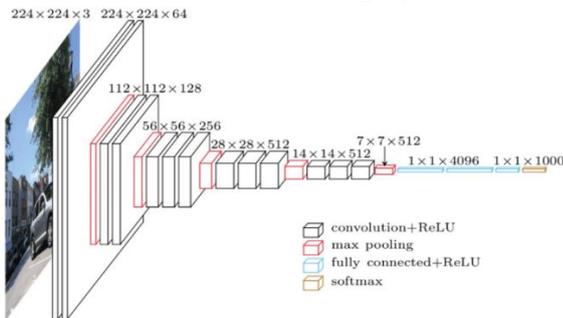

VGG-16 also uses the ImageNet database to train the models with input size 224x224x3 and uses ReLU as the activation function because ReLU introduces non-linearity in the network to map complex model for millions of images within ImageNet. By doing so, ReLU is effective in extracting various features of the images to accurately classify them with the correct label.

Besides, the output size of the depends on the number of fully connected layers in the network. For example, if there are 200 neurons, this will represent 200 classes, resulting in the output size to be 200 for the output layer for the SoftMax activation function.

Compression of the original image is required for the input size of VGG-16, which means that the dimension of the trained image is being compressed from original dimension of 1048 x 810 to 224 x 224 dimension, resulting in features points being lost. Lastly, cosine similarity is used because to compare between two images if they are similar, with an average similarity between 81 to 89 percent.

**7.3.3 Resnet50**
Resnet50 [44] belongs to the Resnet (Residual Network) family with 50 hidden layers. It introduces the concept of residual such as residual block that allows skipping of connection between the earlier to later layers.

$$Residual\ Block\ F(x) = g(x) + x \quad (26)$$

In the equation, $F(x)$ represents the output of the residual block, $g(x)$ represents the residual mapping between the current weights and desired output and $x$ represents as inputs to the residual block.

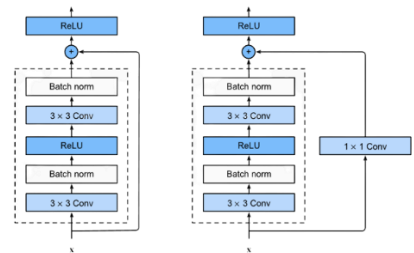

Instead of going through the weigh and activation layers, the residual block will skip to the layers at the end, known a multi-branch inception block. Multi-branch inception block captures information by combining features map from parallel convolution operation. For example, on the right, two 3x3 convolutional filters was used, however it skips these filters by introducing another 1x1 convolutional layer that maps directly to the output to retrieve the residual.

By utilizing the residual connections in ResNet50, the network can perform dimensional reduction by reducing the number of input feature maps. This helps in improving the flow of information through the network and enables more efficient learning by avoiding the degradation of performance that can occur in very deep networks.

VGG-16, Resnet50 also reduces the dimension of the image size to 224 x 224 with the following same steps of feature extraction and cosine similarity with an average similarity value of between 84 to 87 percent.

**7.3.4 Results Comparison**
Below will compare the results of the selected best trained and highest similarity image from the deep Q networks.

**Selected Training Image**

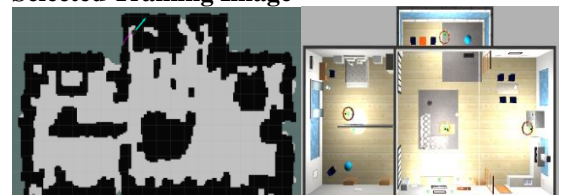

*Fig 70. Trained Image (Left) and AWS House (Right)*

In Figure 70, manual selection of the best trained completed map was selected from the training results that best represents the aws_house environment. As shown, the above trained completed map was chosen as most of the obstacles and outlines are highlighted in the environment.

**Inception V3**
DQN (81% similarity)
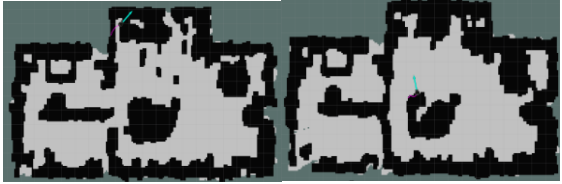

DDQN (80% similarity)
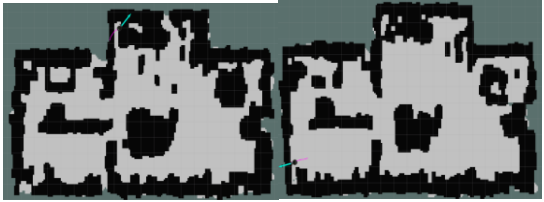

Dueling DQN (80% similarity)
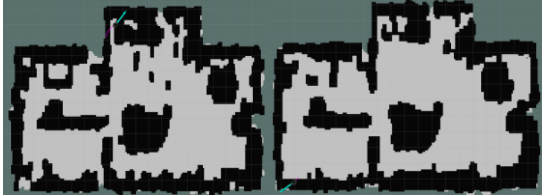

Dueling DDQN (81% similarity)
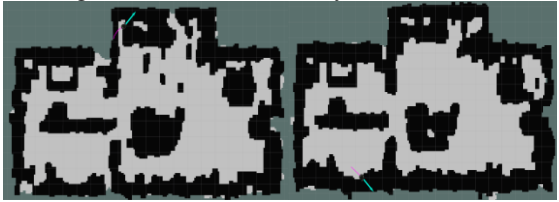

**VGG-16**
DQN (89% similarity)
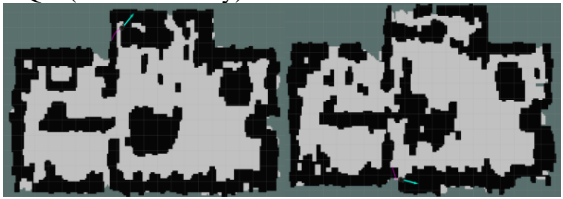

DDQN (87% similarity)
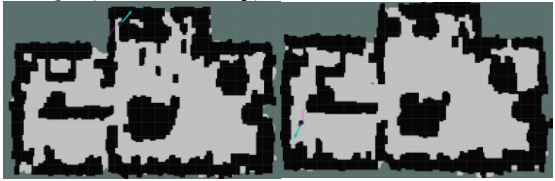

Dueling DQN (88% similarity)
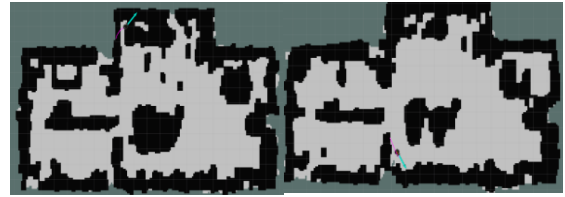

Dueling DDQN (89% similarity)
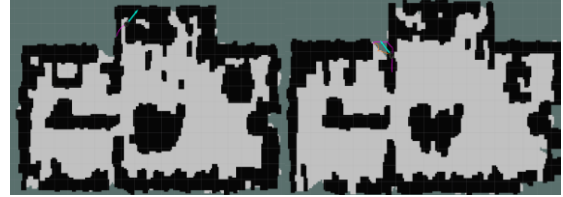

**Resnet50**
DQN (90% similarity)
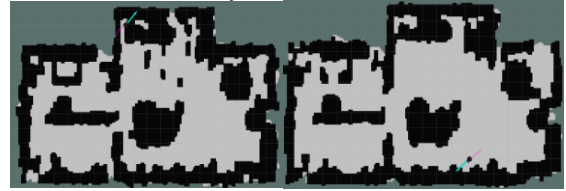

DDQN (89% similarity)
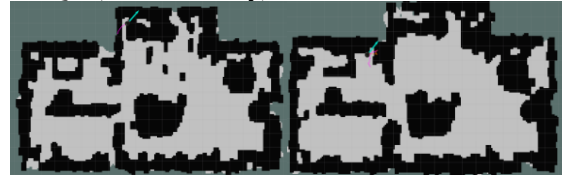

Dueling DQN (89% similarity)
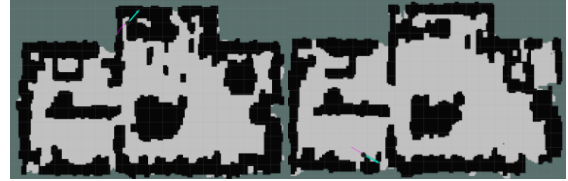

Dueling DDQN (89% similarity)
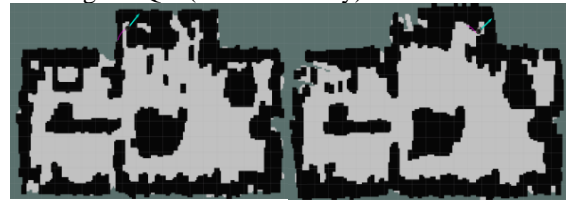

**Conclusion**

In conclusion, based on the evaluation of completed maps, InceptionV3 was chosen as the preferred method for comparison. InceptionV3 utilizes different convolutional filters, including 1x1, 3x3, and 5x5, effectively capturing key features between the trained and generated model maps from the DQN models. This is evident in the similarity scores, where InceptionV3 achieves an average similarity score of 75 to 81 percent, while VGG-16 and ResNet50 achieve scores of 81 to 89 percent.

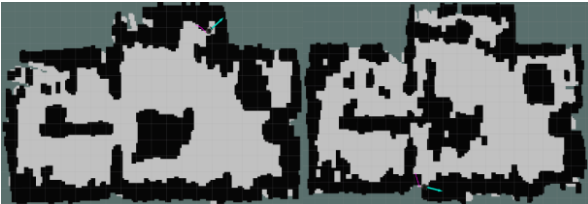

However, when evaluating the completed maps of VGG-16 and Resnet 50, it does not reflect any improvements from the training image because the areas such as the edges and outline of the map have not been properly mapped out.

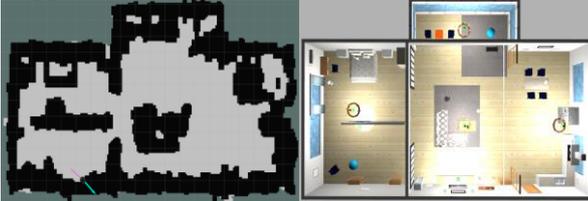

In Figure 84, InceptionV3 successfully maps the entire area of the aws_house environment in all the models where the outline of the edges of various obstacles such as walls and chairs are accurately captured. Therefore, InceptionV3 is chosen as the preferred method for comparison due to its lower similarity value with a better coverage of the completed results of aws_house environment as compared to VGG-16 and Resnet50.

## 8. CONCLUSION

In conclusion, firstly, after conducting various experiments across different various DQN models, Dueling DDQN has emerged as the most effective deep learning reinforcement model for the aws_house Gazebo environment. It demonstrated significant enhancements in key metrics, such as reduced exploration time and an high completion rate of completed map compared to its counterparts. For instance, the exploration time witnessed a noticeable decrease from 700 seconds to 400 seconds, constituting an improvement of approximately 40%. This came along with an increased completion rate across all sets of 5, 10, 15, and 20 experiments. Additionally, when comparing the completed maps between other DQN models using InceptionV3, Dueling DDQN also provides the best results as it captures most of the features and outline of the aws_house environment.

Besides, Dueling DDQN uses a dual-stream architecture composed of estimated state-value $V(s)$ and estimated action-value $Q(s,a)$ after computing the Q-values using DDQN. In the state-value function, it computes the total estimated returns from the state for both immediate and future rewards. In the action-value function, it will return expected return when selecting the action. In the action advantage function $A(s,a)$, it measures the advantage of a selection action in each state. For example, the state-value function estimates the expected return for a state, considering all possible actions according to the policy to determine the best centroid based on the highest information gain. The action-value function estimates the expected return for a specific action taken in a specific state that return the rewards for selecting an action in a particular state with a reward of 0 or 1.

Followed by action advantage function then tells how much better, or a particular action is compared to the average action in that state.

Furthermore, Dueling DDQN uses the architecture of DDQN (Double Q-Learning) to reduce the overestimation of bias that can lead to suboptimal policy decisions. In DDQN, action selection and Q-values generation are decoupled where the model uses two sets of weights, one set to determine the best action to take for a given state for the Q network and the other set to estimate the Q-value of taking that action at the target network. For example, in DDQN, uses a "dones" parameter to ensure that all states are fully completed in the terminal state before going to the next state. Hence, the model can learn more reliably from its past experiences, making it better suited for complex tasks and environments.

Lastly, all results were conducted on a GeForce RTX3060, with maximum and minimum threshold of 450 and 150 with frustum distance of 3 meters. By increasing the frustum distance, additional computational processing will be required due to additional map points detected, which will affect metrics such as exploration time and completion rate depending on the hardware specifications used.

## 9. LIMITATIONS

Firstly, in the current implementation of ExploreORB, this exploration algorithm only works in environments that are small and contain many open areas such as the aws_house environment. For example, in the aws_house environment, there are lesser narrow areas as compared to aws_bookstore such as multiple bookshelves in between the robots. To facilitate the effectiveness of DQN models, it is essential for the robot to map a significant portion of the environment during training when utilizing active SLAM methods. This is crucial because deep reinforcement learning models rely on a comprehensive understanding of the environment to learn and make informed decisions by mapping most of the environment, the DQN model can capture a wide range of scenarios and optimize its learning process, resulting in a more robust and knowledgeable agent.

Besides, currently the number of output size depends on the environment it is exploring. For example, in aws_house, a maximum of 10 possible centroids can be detected at a single time because the environment is smaller. However, for larger environment more centroids will be detected, which results in the modifications of the output size of the environment. Hence, multiple testing is required to find the optimal output size depending on the size and type of environment.

Lastly, to identify the ideal completed map, various CNN models are deployed to help capture similarity between the images. However, if the environment is dynamic, the following comparison methods is not optimal because it is comparing under the assumption that the environment are static

Page 21